\documentclass[letterpaper, 10 pt, conference]{ieeeconf}  

\IEEEoverridecommandlockouts                              

\title{\LARGE \bf
Reinforcement Learning-based Robust Wall Climbing Locomotion Controller in Ferromagnetic Environment
}

\author{Yong Um$^{1,2}$, Young-Ha Shin$^{2}$, Joon-Ha Kim$^{2}$, Soonpyo Kwon$^{1,2}$, and Hae-Won Park$^{1}$
\thanks{$^{1}$Korea Advanced Institute of Science and Technology, Yuseong gu, Daejeon 34141, Republic of Korea. 
haewonpark@kaist.ac.kr}
\thanks{$^{2}$DIDEN Robotics,  Seongdong gu, Seoul 04799, Republic of Korea.}}

\usepackage{amsmath}
\usepackage{amssymb}
\usepackage{mathtools}
\usepackage{booktabs}
\usepackage{tabularx}
\usepackage{cite}
\usepackage{multirow}
\let\labelindent\relax
\usepackage{enumitem}

\begin{document}

\maketitle
\thispagestyle{empty}
\pagestyle{empty}

\begin{abstract}
We present a reinforcement learning framework for quadrupedal wall-climbing locomotion that explicitly addresses uncertainty in magnetic foot adhesion. A physics-based adhesion model of a quadrupedal magnetic climbing robot is incorporated into simulation to capture partial contact, air-gap sensitivity, and probabilistic attachment failures. To stabilize learning and enable reliable transfer, we design a three-phase curriculum: (1) acquire a crawl gait on flat ground without adhesion, (2) gradually rotate the gravity vector to vertical while activating the adhesion model, and (3) inject stochastic adhesion failures to encourage slip recovery. The learned policy achieves a high success rate, strong adhesion retention, and rapid recovery from detachment in simulation under degraded adhesion. Compared with a model predictive control (MPC) baseline that assumes perfect adhesion, our controller maintains locomotion when attachment is intermittently lost. Hardware experiments with the untethered robot further confirm robust vertical crawling on steel surfaces, maintaining stability despite transient misalignment and incomplete attachment. These results show that combining curriculum learning with realistic adhesion modeling provides a resilient sim-to-real framework for magnetic climbing robots in complex environments.
\end{abstract}

\section{INTRODUCTION}

Climbing robots are a promising solution for inspection and maintenance of large-scale steel infrastructure, where manual operations are costly and hazardous ~\cite{hirose1991machine, kim2005spinybotii,autumn2005robotics, asbeck2006scaling,saunders2006rise,bretl2006motion,kennedy2006lemur,brockmann2006concept, kim2008smooth, ward2012design, parness2017lemur, de2018inverted,bandyopadhyay2018magneto, hong2022agile, leuthard2024magnecko}. A variety of adhesion mechanisms have been explored, including suction, gecko-inspired dry adhesion, and magnetic attachment. Among them, magnetic adhesion is particularly well-suited for steel structures, offering strong and repeatable attachment with minimal energy cost, while remaining robust to coatings, dust, and moderate surface irregularities. Combined with the versatility of legged locomotion, magnetic climbing legged robots can traverse complex geometries, step over obstacles, and maintain redundant points of contact for stability, making them attractive for deployment in industrial environments~\cite{bandyopadhyay2018magneto, hong2022agile, leuthard2024magnecko}.

Control of such platforms has primarily relied on Model Predictive Control (MPC)~\cite{hong2022agile,leuthard2024magnecko}, which optimizes motion over a finite horizon under dynamics and kinematic constraints. By incorporating nominal adhesion forces, MPC has enabled stable crawling and trotting on vertical surfaces. However, this reliance on accurate models exposes limitations. First, MPC typically assumes perfect adhesion; in practice, paint, dust, or partial foot placement often reduce holding forces or cause detachment~\cite{hong2022agile}. Since adhesion uncertainty is not modeled, such disturbances often lead to slippage or falls. Second, solving constrained optimizations at every cycle incurs a high computational cost, limiting responsiveness to sudden slip events. Third, mismatches between modeled and actual adhesion forces can destabilize the controller, reducing robustness in unstructured environments.
\begin{figure}[!t] 
    \centering
    \includegraphics[width=0.45\textwidth]{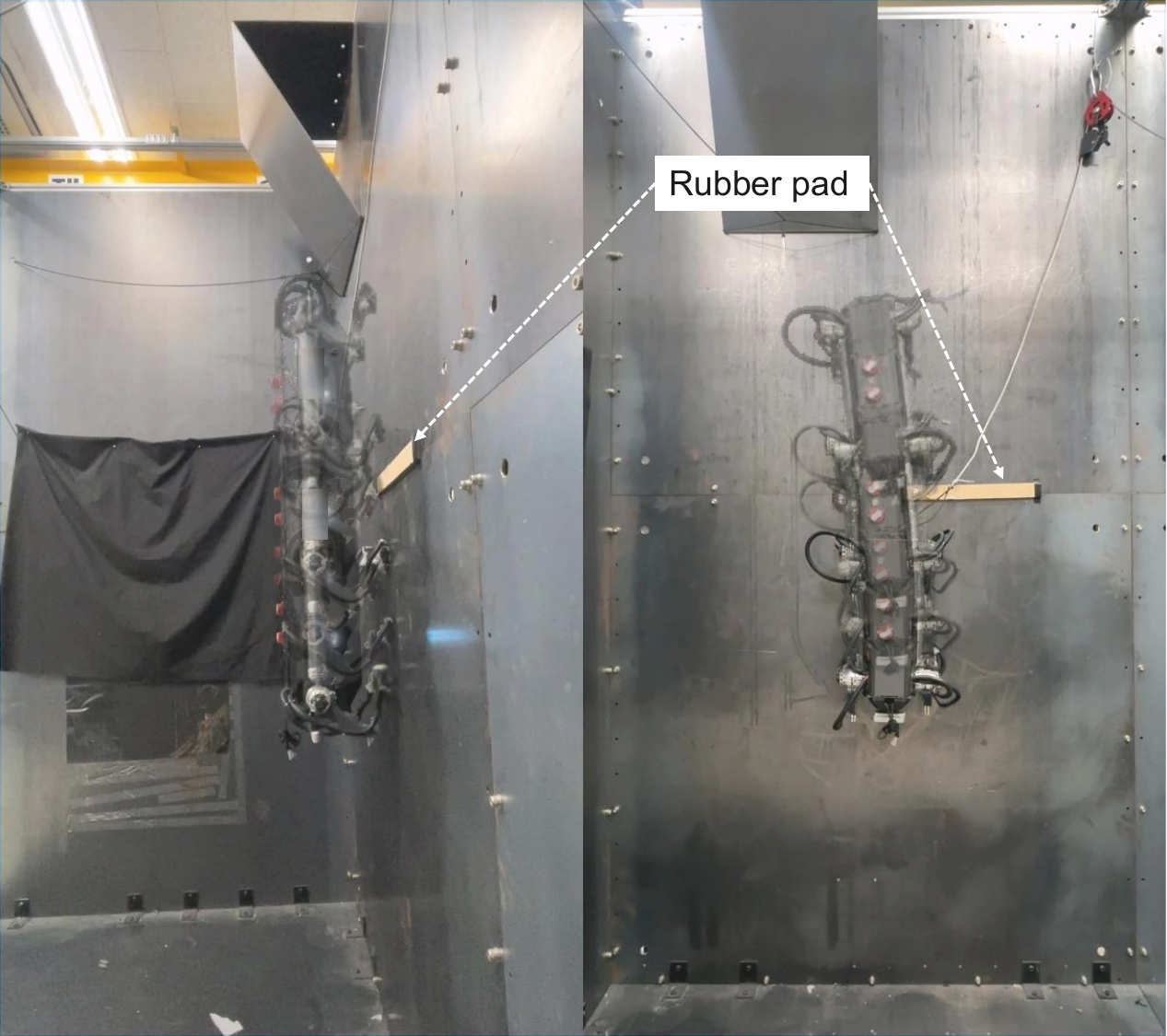}
    \caption{Snapshots of performing robust vertical climbing. 
The learned controller maintains adhesion and recovers from slips even with 
non-ferromagnetic patches, enabling the robot to continue crawling without 
detaching from the wall.}
    \label{fig:main_RL_snapshot}
\end{figure}
Reinforcement learning (RL) offers a complementary paradigm by enabling robots to acquire control policies directly through interaction with their environment~\cite{hwangbo2019learning, lee2020learning, ji2022concurrent, choi2023learning, 10161144, youm2023imitating, miki2022learning, kim2024learning, shin2024reinforcement}. RL-trained controllers have demonstrated robust locomotion over rough terrains and recovery from significant perturbations, often exceeding the adaptability of model-based approaches. Extending RL to vertical climbing is particularly promising, since an RL policy could, in principle, learn recovery from adhesion failures without explicit adhesion models. Yet, climbing introduces unique challenges: adhesion must be accurately represented in simulation to avoid a sim-to-real gap, and adhesion loss often leads to catastrophic falls. Moreover, naive RL training frequently produces policies that fail to generalize to vertical settings. Prior work has rarely addressed RL-based climbing, and almost none have explicitly considered adhesion uncertainty during training.

In this work, we propose a reinforcement learning framework for a quadrupedal magnetic climbing robot~\cite{hong2022agile}. Our contributions are threefold:  
(1) a multi-phase curriculum that gradually rotates the gravity vector from horizontal to vertical for smooth transition from ground crawling to wall climbing;  
(2) a physics-based adhesion model of electropermanent-magnetic (EPM) feet that captures both partial and failed contacts; and  
(3) stochastic adhesion failures during training to encourage recovery under uncertainty.  

Simulation and hardware experiments show that our learned controller achieves robust vertical climbing, maintains adhesion despite failures, and recovers from slip events, whereas an MPC baseline fails immediately. These results highlight the importance of curriculum learning and realistic adhesion modeling for reliable sim-to-real transfer in magnetic climbing robots Fig.~\ref{fig:main_RL_snapshot}.

\section{METHODS}
%
\noindent
\label{METHODS}
\begin{figure*}[t]
    \centering
    \includegraphics[width=1.0\textwidth]{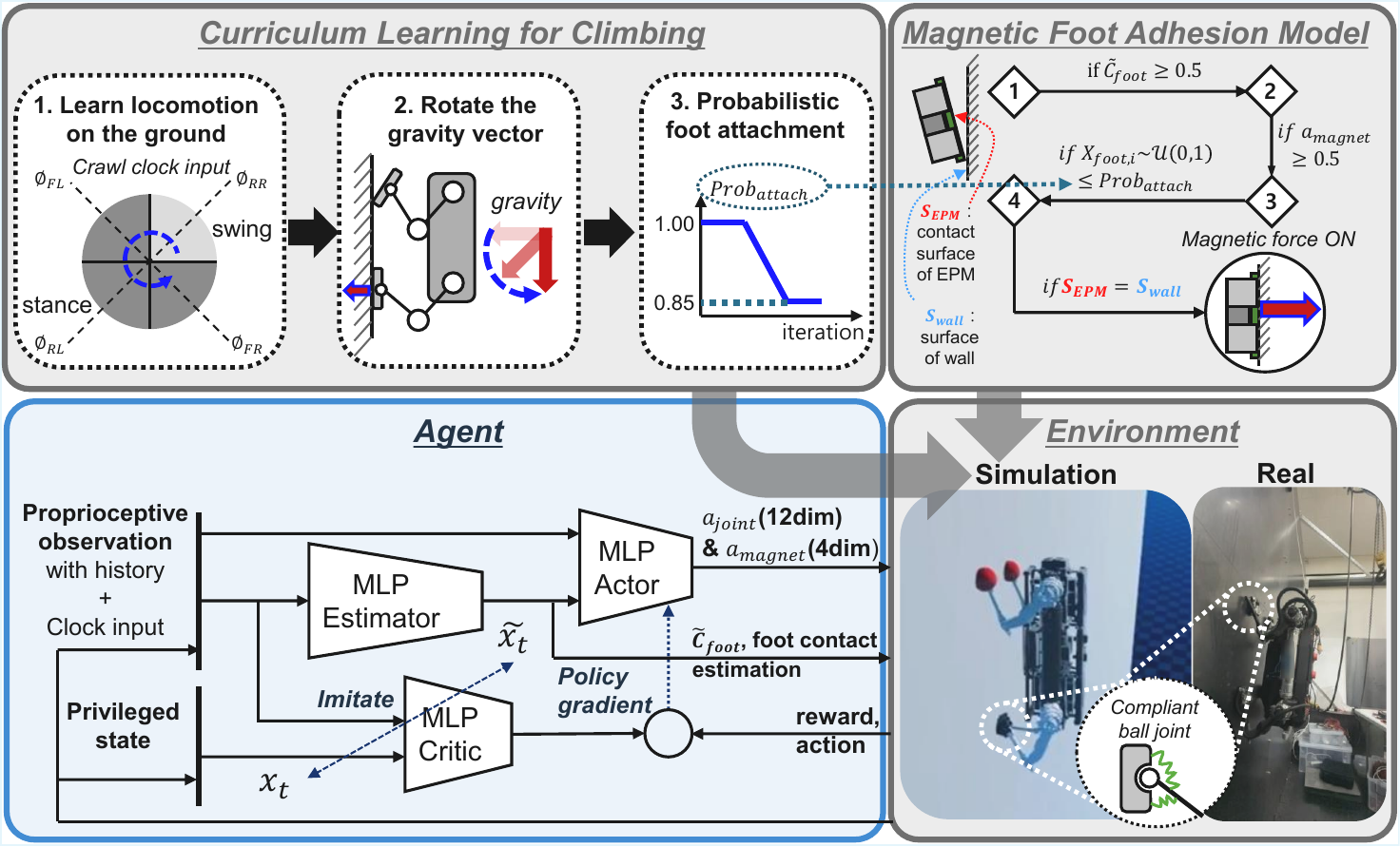}
    \caption{Overview of the proposed learning framework for vertical locomotion under adhesion uncertainty. 
    The process consists of four main components: (1) ground locomotion pre-training, (2) curriculum-based gravity rotation to adapt to climbing orientation, (3) probabilistic foot adhesion modeling to simulate imperfect magnetic contact, and (4) integration of simulation-to-real transfer with a compliant magnetic foot model. The policy receives proprioceptive observations with history and a crawl clock input, while an estimator predicts foot contacts. The actor--critic architecture is trained via reinforcement learning and imitation signals to produce joint torques and magnetic adhesion actions.}
    \label{fig:Learning_Framework_Overview}
\end{figure*}
We propose a reinforcement learning–based control framework for robust vertical crawling on steel surfaces with uncertain magnetic adhesion. A realistic magnetic foot adhesion model is incorporated into the simulation to reflect the behavior of electropermanent-magnetic feet. The training process follows a curriculum that gradually rotates the gravity vector from horizontal to vertical, enabling the policy to adapt to climbing conditions. To improve robustness, we introduce stochastic foot adhesion failures during training. An overview of the overall training and deployment framework is illustrated in Fig.~\ref{fig:Learning_Framework_Overview}, and the learned policy is deployed on the quadrupedal magnetic climbing to demonstrate stable vertical crawling in real-world experiments.

\subsection{Hardware}
\noindent
\label{Hardware}
We utilize an untethered quadrupedal climbing robot designed for dynamic locomotion on vertical steel surfaces. The robot weighs 8 kg and is equipped with four magnetic feet, each weighing approximately 0.2 kg, integrated with electropermanent magnets (EPMs) that provide switchable magnetic adhesion \cite{hong2022agile}.
The EPMs used in this study generate a maximum normal holding force of approximately 697 N and support rapid magnetic switching within 5 milliseconds, allowing timely attachment and detachment during locomotion. While prior designs incorporated magnetorheological elastomer (MRE) footpads to enhance shear adhesion, we employ a simplified foot design without MRE to better isolate and model adhesion uncertainty during learning.
\subsection{Magnetic Foot Adhesion Model}

\begin{figure}[!t] 
    \centering
    \includegraphics[width=0.4\textwidth]{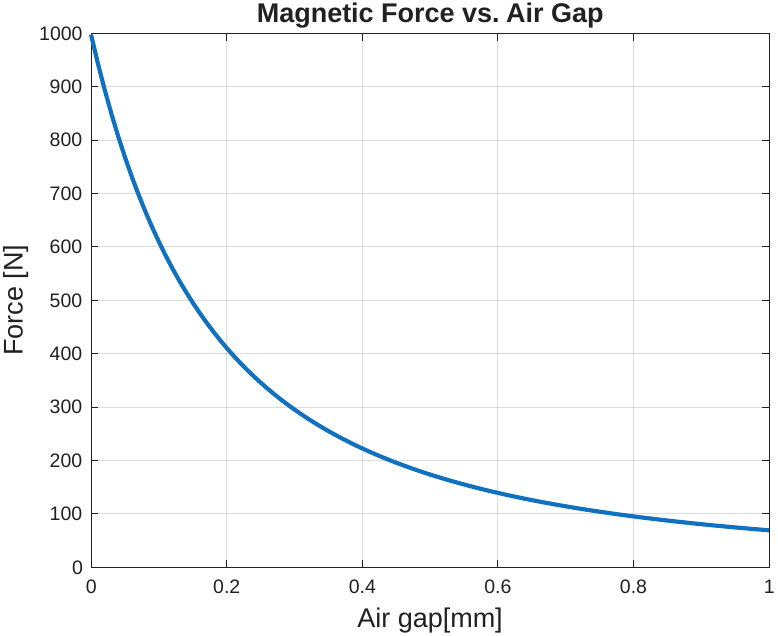}
    \vspace{-1mm}
    \caption{Magnetic force of the magnetic feet with respect to air gap.
    Even with a 1 mm air gap, the adhesion force drops to about 7\% of the maximum value.}
    \label{fig:Air gap magnetic force}
\end{figure}

\noindent
\label{Magnetic Foot Adhesion Model}
To realistically capture the behavior of electropermanent-magnetic (EPM) feet, 
we implement a magnetic adhesion model in simulation. 
EPMs generate holding force by magnetizing an AlNiCo core through a coil pulse, which requires forming a closed magnetic circuit with the steel surface. 
Stable adhesion is obtained only when the EPM is in proper contact with the surface; if the magnet is activated without contact, 
or if an air gap exists between the EPM and the plate, the resulting force is significantly reduced. 
Fig.~\ref{fig:Air gap magnetic force} illustrates how the adhesion force drops as the gap increases, 
highlighting the sensitivity of EPMs to partial contact conditions.

In simulation, adhesion is considered only if the following conditions are satisfied in sequence:

\textbf{(1) Contact recognition.} 
The state estimator outputs a contact confidence $\tilde{c}_{\text{foot}} \in [0,1]$, 
which is interpreted as the probability of foot–wall contact. 
A contact is recognized when $\tilde{c}_{\text{foot}} \geq 0.5$, 
corresponding to the binary contact indicator $c_{\text{foot}} = 1$ (contact) and $c_{\text{foot}} = 0$ (no contact):
\begin{equation}
\label{eq:contact_confidence}
\tilde{c}_{\text{foot}} \geq 0.5.
\end{equation}

\textbf{(2) Magnet activation.} 
The policy outputs an activation signal $a_{\text{magnet}}$ trained to regress the binary contact indicator $c_{foot} \in \{0,1\}$. 
Since $c_{foot}$ takes only 0 or 1, we use 0.5 as a threshold: 
\begin{equation}
\label{eq:magnet_activation}
a_{\text{magnet}} \geq 0.5,
\end{equation}
meaning that the magnet is regarded as ON when the output exceeds this value.

\textbf{(3) Stochastic adhesion.}  
Even when both conditions hold, adhesion occurs only probabilistically to model real-world uncertainties. 
A random variable $X \sim \mathcal{U}(0,1)$ is sampled only when the foot is in swing (not contact with the wall), and adhesion succeeds only if
\begin{equation}
\label{eq:stochastic_adhesion}
 X \leq \text{Prob}_{\text{attach}}.
\end{equation}
This stochastic component represents environment-dependent failures such as non-ferrous surfaces, uneven or painted walls, or partial contacts. 
The details of $\text{Prob}_{\text{attach}}$ scheduling are given in Section~\ref{subsec:Multi-Phase Learning Strategy for Vertical Locomotion under Adhesion Uncertainty}.

\textbf{(4) Geometric alignment.}  
Finally, proper adhesion requires full geometric contact between the magnet and the wall surface:
\begin{equation}
\label{eq:contact_condition}
S_{\text{EPM}} = S_{\text{wall}}.
\end{equation}
To generate a stable adhesion force, the EPM surface must be completely aligned with the wall to form a closed magnetic circuit.
In cases where even a small gap remains, the magnetic circuit is incomplete and the effective adhesion force is drastically reduced.
By enforcing these sequential conditions, the policy is exposed to realistic adhesion failures 
and learns robust recovery strategies that transfer effectively to real hardware. 
This modeling choice reduces the sim-to-real gap by capturing imperfect contact and slippage, 
enabling reinforcement learning to produce policies robust to unreliable adhesion. 
Further details and empirical analyses of this strategy are presented in Section~\ref{sec:results}.
In hardware implementation, the electropermanent magnet (EPM) is controlled via current pulses. 
When the contact recognition (Eq.~\ref{eq:contact_confidence}) and magnet activation (Eq.~\ref{eq:magnet_activation}) conditions are satisfied, a current pulse is applied to switch the EPM \emph{on} and generate adhesion force. 
Conversely, when these conditions become violated, a reverse current pulse is applied 
to switch the EPM \emph{off}, thereby releasing adhesion.

\subsection{Multi-Phase Learning Strategy for Vertical Locomotion under Adhesion Uncertainty}
\label{subsec:Multi-Phase Learning Strategy for Vertical Locomotion under Adhesion Uncertainty}

To enable robust and adaptive climbing locomotion, we adopt a multi-phase learning strategy that progressively advances from basic gait learning to climbing-specific challenges. As shown in Fig.~\ref{fig:Learning_Framework_Overview}, the training is organized into three sequential phases: (1) gait acquisition on flat terrain, where the robot first learns a stable crawl without adhesion to avoid frequent early failures, (2) adaptation to altered gravitational orientations during wall climbing, and (3) robustness learning under probabilistic foot adhesion failures that mimic imperfect magnetic contact in real hardware.

\paragraph{Phase 1: Gait Acquisition on Flat Ground}

The first phase aims to establish a stable crawl gait on flat ground before introducing gravitational transitions or adhesion uncertainty. We begin training on flat ground because directly attempting to learn wall climbing from scratch leads to frequent falls: Untrained initial policies, generated from random initialization, are unable to reliably lift and place their feet while maintaining body support. As a result, they frequently fall off the wall during early training, which prevents the collection of sufficient high-quality samples for stable learning. On flat ground, by contrast, the robot can safely acquire a baseline crawling gait without magnetic adhesion. During this stage, the policy is allowed to output magnetic foot actions, but the adhesion model is deliberately disabled so that no actual adhesive force is applied. This setup ensures that the controller first concentrates on learning smooth and coordinated crawling, without relying on adhesion. At the same time, an auxiliary reward is introduced to guide when magnets should ideally be switched on or off: it penalizes the policy if magnet activation occurs during swing or if magnets are off during stance. Through this mechanism, the policy learns a natural timing of adhesion—turning magnets ON during stance and OFF during swing—even before real adhesion forces are simulated. This implicit learning of magnet timing provides a foundation for stable climbing behaviors, which will be leveraged in later phases when adhesion is physically modeled. Training follows the concurrent learning approach in~\cite{ji2022concurrent}, with the addition of an 8-dimensional clock input to encode each leg’s gait phase and promote periodic behaviors such as crawling.

\paragraph{Phase 2: Adaptation to Gravitational Rotation}

After learning locomotion on flat ground, the robot is gradually exposed to inclined and vertical surfaces by rotating the direction of gravity in the simulation. This curriculum-based gravitational transition enables the policy to adapt its balance and posture control to non-horizontal terrains. During this phase, the magnetic adhesion model is activated, and foot contact becomes critical for maintaining stability on climbing surfaces. The timing strategy for magnet activation, which was shaped in Phase~1 through auxiliary rewards, is now coupled with actual adhesion forces so that the robot can reliably attach its feet during stance and release them during swing.

The gravity rotation is scheduled per training iteration $t$ as:
\begin{equation}
\label{eq:theta}
\theta(t) = \min\!\left\{ \tfrac{\pi}{2},\; \max\!\left\{ 0,\; \tfrac{\pi}{2}\cdot \tfrac{t-1200}{20000} \right\} \right\},
\end{equation}
where $\theta(t)$ is the tilt angle from the ground ($0^\circ$) to the wall ($90^\circ$). 
Let $g_0 \in \mathbb{R}^3$ denote the nominal gravity vector in the ground frame. 
To gradually shift the environment from flat ground to a vertical wall, 
$g_0$ is rotated by $\theta(t)$ around an axis parallel to the ground surface. 
In this study, gravity is defined along the $-z$ direction and rotated about the $+y$ axis, yielding
\begin{equation}
g(t) = R_y\!\big(\theta(t)\big)\, g_0 , \qquad R_y(\theta) \in \mathrm{SO}(3).
\end{equation}

Thus, for $t \le 1200$, the robot trains on flat ground with magnetic forces already applied; from $t=1201$ to $t=21200$, the gravity vector is linearly tilted until it reaches $90^\circ$; and for $t > 21200$, the gravity vector remains at $90^\circ$ for the remainder of training.

\paragraph{Phase 3: Robustness to Adhesion Uncertainty}

In the final phase, we introduce stochastic disturbances to simulate real-world adhesion uncertainty. Even when a magnetic foot is commanded ON, the attachment may fail with a certain probability, reflecting realistic failure modes such as partial foot placement, non-ferromagnetic surface coatings, or surface contamination. By exposing the policy to these perturbations during training, the robot learns to maintain balance and recover from unexpected slippage or detachment.

The adhesion success probability $\text{Prob}_{\text{attach}}(t)$ is kept at $1.0$ until iteration $t=21200$ and then decreased linearly to $0.85$ by iteration $t=35000$:
\begin{equation}
\label{eq:prob_attach}
\text{Prob}_{\text{attach}}(t) 
= 1.0 - 0.15 \cdot 
\frac{\min\!\big(\max(t-21200,\,0),\,13800\big)}{13800}.
\end{equation}
During training, adhesion is not triggered solely by stance phase, but only when all activation conditions are satisfied:  the contact estimator must indicate a valid foot contact ($\tilde{c}_{\text{foot}} \geq 0.5$, Eq.~\ref{eq:contact_confidence}), the commanded magnet action must exceed the activation threshold ($a_{\text{magnet}} \geq 0.5$, Eq.~\ref{eq:magnet_activation}), and  the foot must be in contact with the wall surface ($S_{\text{EPM}} = S_{\text{wall}}$, Eq.~\ref{eq:contact_condition}). Once these conditions hold, the adhesion model samples a random number uniformly from $[0,1]$ and compares it with the scheduled success probability $\text{Prob}_{\text{attach}}(t)$ (Eq.~\ref{eq:stochastic_adhesion}). If the sample is smaller, the foot attaches successfully and produces magnetic force; otherwise, the attempt fails and no holding force is applied. This mechanism injects probabilistic failures into training, encouraging the policy to learn recovery strategies that maintain stability even under imperfect adhesion.

\subsection{Learning Framework}

We adopt a reinforcement learning framework based on Proximal Policy Optimization (PPO) to train a robust climbing locomotion policy in the RaiSim simulator~\cite{hwangbo2018per}. 
The framework consists of three neural networks: an actor (policy) network, a critic (value function) network, and a state estimator network that predicts privileged states such as base velocity, foot height, and foot contact probabilities~\cite{ji2022concurrent}. 
All networks are implemented as multilayer perceptrons (MLPs). 
The actor and critic share the same architecture with three hidden layers of sizes $[256,\,128,\,64]$, while the state estimator is a smaller MLP with two hidden layers of sizes $[256,\,128]$. 
The overall structure is illustrated in Fig.~\ref{fig:Learning_Framework_Overview}.
\paragraph{Observation and Action Space}
The observation vector $o_t$ aggregates proprioceptive states, historical information, 
and gait phase encoding. Specifically, it contains joint positions $q_t$ and velocities $\dot{q}_t$; 
previous joint position targets for the last two time steps; base orientation $\phi_t$ and angular velocity $\omega_t$; 
Cartesian foot positions relative to the base; estimated base linear velocity, foot height, and foot contact probabilities from the state estimator; 
and an 8-dimensional clock input $o_{\text{clock}}$ for gait phase encoding.
The clock input is defined as 
$o_{\text{clock}} = [\sin \phi_i, \cos \phi_i]_{i=1}^4 \in \mathbb{R}^8$, 
where $i \in \{1,2,3,4\}$ corresponds to the right-rear (RR), right-front (FR), left-rear (RL), and left-front (FL) legs. 
The phase variables are given by $\phi_i = \tfrac{2\pi}{T}t + \tfrac{\pi}{2}i$, 
where $t$ is the time index, $T=1.2\,\text{s}$ denotes the gait cycle period, 
and $\tfrac{\pi}{2}i$ introduces a quarter-phase shift between consecutive legs. 
All observation signals are further processed through a first-order low-pass filter 
before being fed into both the actor and the estimator networks. 
The filter is implemented as $obs_{\text{filter}} = (1-\alpha)obs_{\text{old}} + \alpha obs_{\text{new}}$, 
with the smoothing coefficient set to $\alpha=0.35$. 
This filtering suppresses high-frequency noise while retaining essential state information 
for stable policy learning.

\paragraph{Reward Design}
The reward function is designed to produce stable, efficient, and robust climbing behaviors while discouraging unsafe or inefficient actions. It is composed of multiple components that address different aspects of locomotion, such as velocity tracking, posture stability, foot placement, and action smoothness. Table ~\ref{tab:reward_terms} lists all reward terms and their mathematical expressions.
To facilitate curriculum learning in the climbing task, several rewards are dynamically scaled over training iterations. 
We define the scheduling factor $\kappa = 0.99975^{\max(iter-1200,0)}$. The velocity tracking rewards ($R_{lv}, R_{av}$) are scaled by $(1.5-0.5\kappa)$, 
which gradually increases their weight to emphasize accurate command following when the robot transitions to vertical climbing. 
Conversely, the foot slip ($R_{fs}$) and joint torque ($R_{\tau}$) penalties are scaled by $(0.5+0.5\kappa)$, 
which gradually decreases their weight, allowing more torque usage and occasional foot slip during wall climbing without overly penalizing the agent.
Furthermore, the action smoothness terms ($R_{as1}, R_{as2}$) are disabled during the early stage of Phase~1 ($t<1000$) 
to prevent over-constraining exploration. After this point, they are applied as described in Table~\ref{tab:reward_terms}.

\begin{table}[!htbp] 
\caption{Reward components used in the climbing locomotion task.}
\label{tab:reward_terms}
\vspace{-2mm} 
\centering
\renewcommand{\arraystretch}{1.2}
\setlength{\tabcolsep}{6pt}
\begin{tabularx}{\linewidth}{lX}
\toprule
\textbf{Reward} & \textbf{Expression / Description} \\
\midrule
Linear velocity ($R_{lv}$) &
$\big(1.5-0.5\,\kappa\big)\;
3.0\,\exp\!\big(-5.0\,\|v^{\text{desired}}_{xy}-v_{xy}\|^2\big)$ \\

Angular velocity ($R_{av}$) &
$\big(1.5-0.5\,\kappa\big)\;
3.0\,\exp\!\big(-5.0\,(\omega^{\text{desired}}_{z}-\omega_{z})^2\big)$ \\

Standing command ($R_{sc}$) &
$\displaystyle 0.5 \sum_{i=1}^{4} r_i$ \\

Gait ($R_g$) &
$\displaystyle 0.5\sum_{i=1}^{4} g_i \quad$ \\

Foot height ($R_{fh}$) &
$\displaystyle 0.5\,\exp\!\big(-\sum_i f_i\,(p^{\text{des}}_{z,i}-p_{z,i})^2)$ \\

Foot slip ($R_{fs}$) &
$\big(0.5+0.5\,\kappa\big)\;
0.5\sum_i c_i\,\|v_{xy,i}\|^2$ \\

Foot clearance ($R_{fc}$) &
$\displaystyle 
140 \sum_i (1\!-\!c_i)\,(p^{\text{des}}_{z,i}-p_{z,i})^2 \,\|v_{z,i}\|^{0.5}$ \\

Orientation ($R_o$) &
$\displaystyle 3\cdot \text{angle}\!\big(\phi_{\text{body},z},\,v_{\text{world},z}\big)$ \\

Joint torque ($R_{\tau}$) &
$\big(0.5+0.5\,\kappa\big)\;
0.003\,\|\tau_t\|^2$ \\

Joint position ($R_{jp}$) &
$\displaystyle \alpha_{jp}\,\|q_t-q^{\text{nominal}}\|^2$ \\

$\displaystyle 0.003 \| \dot{q} \|^2$       \\ 
Joint acceleration ($R_{ja}$) &
$\displaystyle 0.003\,\|\ddot{q}_t\|^2$ \\
Action smoothness 1 ($R_{as1}$) &
$\displaystyle 
\begin{cases}
0, \\
\text{if Phase 1 and } t<1000, \\[3pt]
2.5\,\|a_{t}-a_{t-1}\|^2, \\
\text{otherwise},
\end{cases}$ \\
Action smoothness 2 ($R_{as2}$) &
$\displaystyle 
\begin{cases}
0, \\
\text{if Phase 1 and } t<1000, \\[3pt]
1.2\,\|a_{t}-2a_{t-1}+a_{t-2}\|^2, \\
\text{otherwise},
\end{cases}$ \\
Base motion ($R_{bm}$) &
$\displaystyle 3.0\,\exp\!\big(-0.5\,\|\omega_{x,y}\|^2+0.2\,|v_z|\big)$ \\
Action magnet ($R_{am}$) &
$\displaystyle 0.15\sum_i \big(c_{i}-a_{magnet,i}\big)^2$ \\
\bottomrule
\end{tabularx}
\vspace{-2mm} 
\end{table}
\noindent\textbf{Notation.}
\begin{gather}
g_i =
\begin{cases}
  1,  & \phi_i\in(0,\tfrac{\pi}{2}),\ \text{foot $i$ not in contact},\\
 -1,  & \phi_i\in(0,\tfrac{\pi}{2}),\ \text{foot $i$ in contact},\\
  1,  & \phi_i\notin(0,\tfrac{\pi}{2}),\ \text{foot $i$ in contact},\\
 -1,  & \text{otherwise},
\end{cases} \\[3pt]
f_i =
\begin{cases}
  1, & \phi_i\in(0,\tfrac{\pi}{2}),\\
  0, & \text{otherwise},
\end{cases} \\[3pt]
p^{\text{des}}_{z,i} =
\begin{cases}
  0.08, & \phi_i\in(0,\tfrac{\pi}{2}),\\
  0,    & \text{otherwise},
\end{cases} \\[3pt]
c_i =
\begin{cases}
  1, & \text{if foot $i$ in contact},\\
  0, & \text{otherwise},
\end{cases} \\[3pt]
r_i =
\begin{cases}
  1,  & \text{if foot $i$ in contact when } v^{\text{desired}}=0,\\
 -1,  & \text{otherwise},
\end{cases} \\[3pt]
\alpha_{jp} =
\begin{cases}
  3,    & \text{if } v^{\text{desired}}=0 \ \text{(standing command)},\\
  0.75, & \text{otherwise}.
\end{cases}
\end{gather}

\noindent
$p_{z,i}$: vertical position of foot $i$; \quad
$v_{xy,i}$: horizontal velocity of foot $i$; \quad
$v_{z,i}$: vertical velocity of foot $i$; \quad
$\phi_i$: gait phase of leg $i$; \quad
$\tau_t$: joint torques; \quad
$q_t$: joint positions; \quad
$\ddot{q}_t$: joint accelerations; \quad
$a_{magnet,i}$: magnet action for leg $i$.

\noindent\textbf{Total reward:}
\begin{multline}
\small
R_{\text{tot}} = \big(R_{lv}+R_{av}+R_{g}+R_{fh}+R_{sc}\big)\cdot \\
\exp\Big\{-0.2\big(R_{fs}+R_{fc}+R_{o}+R_{\tau}+R_{jp}+R_{js}+R_{ja} \\
{}+ R_{as1}+R_{as2}+R_{bm}+R_{am}\big)\Big\}.
\end{multline} 
This formulation promotes forward crawling with correct gait timing while suppressing undesirable behaviors such as excessive slipping, abrupt motions, or unstable body orientation. The exponential term acts as a multiplicative penalty, amplifying the effect of violations in stability, contact, and actuation smoothness.

\paragraph{Domain Randomization}
To ensure that the policy trained in simulation can be reliably deployed on the real robot, we incorporate domain randomization and hardware-aware modeling throughout the training process. This approach mitigates the sim-to-real gap by exposing the policy to a range of environmental and actuation variations during training.
At the beginning of each episode, the following parameters are uniformly randomized within predefined ranges:
\begin{itemize}
    \item \textbf{Joint PD gains:} $[0.4, 0.6]$ to represent joint P gain, and $[0.12, 0.18]$ to represent joint D gain.
    \item \textbf{Ground friction coefficient:} $[0.3, 0.5]$ to represent smooth painted steel and rough uncoated surfaces.
    \item \textbf{Observation noise:} Uniform noise applied to observation channels: 
    orientation ($\pm0.05$~rad plus an offset bias sampled within $\pm0.05$~rad), 
    joint angles ($\pm0.1$~rad), 
    body angular velocity ($\pm0.1$~rad/s), 
    joint velocities ($\pm0.5$~rad/s), 
    joint position histories ($\pm0.1$~rad), 
    joint velocity histories ($\pm0.5$~rad/s), 
    and foot positions ($\pm0.015$~m).
    \item \textbf{Action delay:} Random delay in the range $[0, 0.008]$~s to emulate hardware switching latency.
\end{itemize}

\paragraph{Compliant 3-DOF Joint Ankle Modeling}
To realistically capture the mechanical compliance of the ankle, the calf link (shank) and the magnetic foot are connected via a ball joint with three rotational degrees of freedom. The nominal ankle orientation is fixed as the position target, and elastic compliance is modeled by applying position and velocity gains that mimic the effect of the elastic band connecting the foot to the shank. To improve robustness, these gains are randomized during training within a predefined range. The nominal values of the target angle and the controller gains are summarized below:
\begin{itemize}
    \item Nominal ankle orientation: \\
    roll–pitch–yaw $[0,0.523599,0]~\text{rad}$ ($[0^\circ,30.0^\circ,0^\circ]$), 
    which follows the definition of the nominal configuration reported in~\cite{hong2022agile}.
    \item Position gain ($K_p$): $0.05  \pm 0.01$
    \item Velocity gain ($K_d$): $0.001 \pm 0.0005$
\end{itemize}

\section{RESULTS}
\label{sec:results}
\subsection{Learning Progress Across Phases}

\begin{figure}[t]
    \centering
    \includegraphics[width=\linewidth]{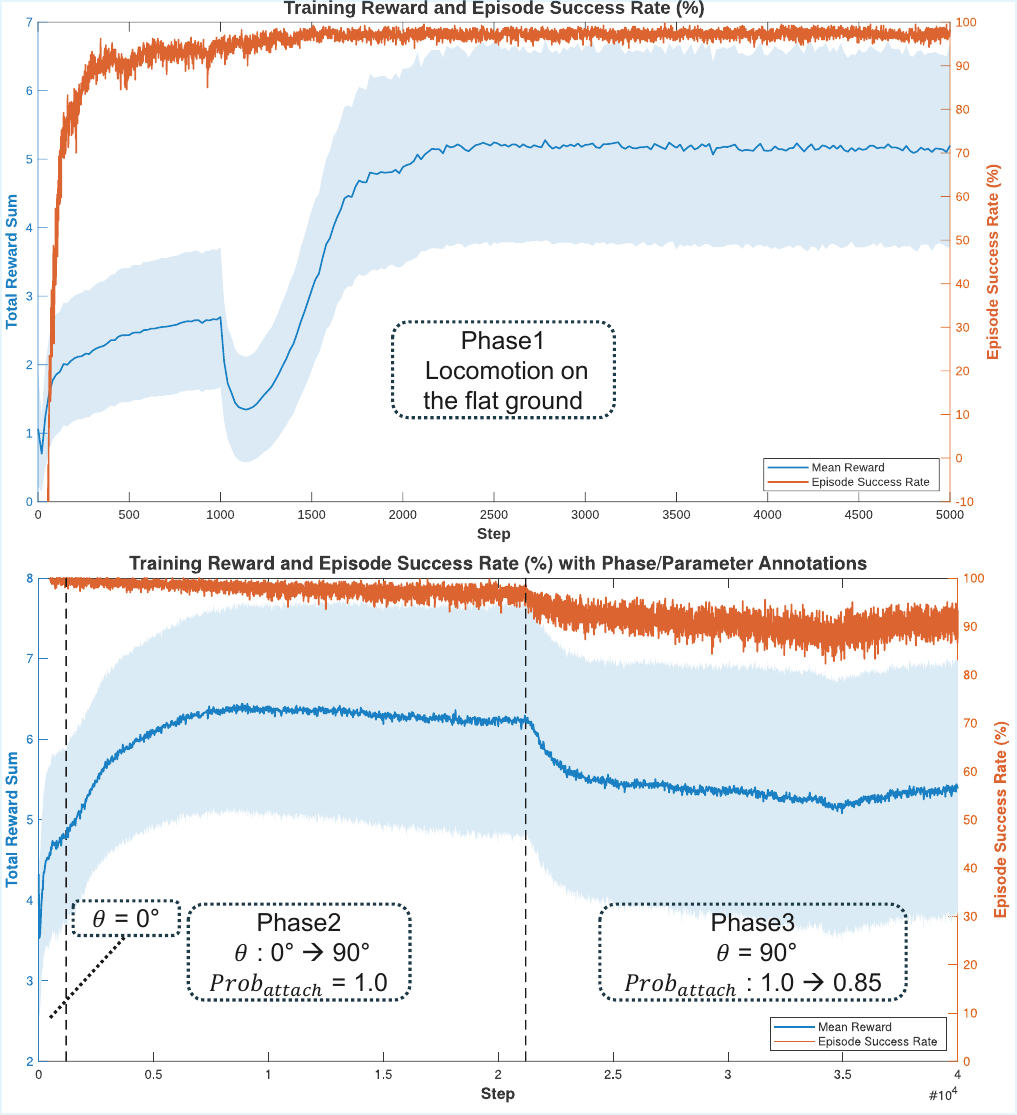}
    \caption{Training curves of climbing policy optimization. 
    The vertical dashed lines indicate phase transitions of the multi-phase learning curriculum. 
    Phase~1 trains locomotion on flat ground. 
    In Phase~2, the wall inclination $\theta$ increases from $0^\circ$ to $90^\circ$ with perfect adhesion ($\text{Prob}_{\text{attach}}=1.0$). 
    Phase~3 fixes $\theta=90^\circ$ while gradually reducing the adhesion probability from $1.0$ to $0.85$, exposing the policy to adhesion failures.}
    \label{fig:learning_curves}
\end{figure}

Fig.~\ref{fig:learning_curves} shows the learning progress of the proposed curriculum.  
During Phase~1, the policy rapidly improves reward as it acquires a stable crawl gait on flat ground. 
After $t>1000$, the action smoothness regularization terms ($R_{as1}, R_{as2}$) are activated, 
which temporarily lowers the total reward due to the additional penalty. 
However, the policy quickly adapts, and the reward continues to increase as smoother and more stable motions emerge.  

During Phase~2, the gradual rotation of the gravity vector toward the wall leads to a reduction in the average episode success rate.  
In Phase~3, the introduction of stochastic adhesion increases the likelihood of detachment from the wall, which in turn induces a gradual decline in overall performance. 
However, the performance eventually saturates, and despite the adhesion probability being reduced to $85\%$, the policy consistently shows an average episode success rate of approximately $90\%$.  
This shows that the curriculum design stabilizes training and allows the policy to be generalized to increasingly challenging climbing conditions.

\subsection{Ablation Study}

\paragraph{Training and Evaluation Settings}
Each ablation policy was trained under identical reinforcement learning settings,
with the only difference being the removed component. All policies were trained
using PPO in RaiSim with proprioceptive observations, clock inputs, and domain randomization.
The reward structure and training duration were kept consistent across all runs.
The following summarizes the conditions for each ablation:
\begin{itemize}
    \item \textbf{Full (ours):} Multi-phase curriculum learning was applied,
    gradually rotating the gravity vector from $0^{\circ}$ to $90^{\circ}$,
    with realistic magnetic foot modeling enabled and stochastic adhesion
    failures introduced according to Eq.~\ref{eq:stochastic_adhesion}.

    \item \textbf{w/o Curriculum:} The gravity vector was fixed at $90^{\circ}$ 
    from the beginning of training, and the reward formulation was identical 
    to that used in Phases~2 and~3. Realistic adhesion modeling and stochastic 
    adhesion failures were still applied.

    \item \textbf{w/o Probabilistic Adhesion:} The probability of foot adhesion
    was fixed to $\text{Prob}_{\text{attach}}=1.0$ during training, disabling
    stochastic failures. Gravity curriculum and realistic modeling remained enabled.

    \item \textbf{w/o Modeling:} Realistic adhesion modeling, including 
    alignment and air-gap effects, was removed. In this setting, adhesion was 
    considered ideal whenever the magnet command($a_{magnet}$) exceeded $0.5$, regardless of the contact geometry. Thus, even partial contact between the EPM and the wall surface was treated as full adhesion. For fair comparison, $\text{Prob}_{\text{attach}}$ was fixed to $1.0$ to isolate the effect of modeling.

\end{itemize}



\paragraph{Evaluation Metrics}
We define the following metrics to evaluate climbing performance, 
using a fixed evaluation horizon of $T_{\mathrm{hor}}=10$\,s and $N=100$ episodes.






\begin{itemize}
    \item \textbf{Velocity Tracking RMSE:} quantifies how accurately the robot follows the commanded velocity profile. Lower values indicate better tracking performance. The velocity command is defined as 
    $v^{\mathrm{desired}}(t) = [v_{x}^{\mathrm{desired}}(t),\, v_{y}^{\mathrm{desired}}(t),\, \omega_{z}^{\mathrm{desired}}(t)] 
    \in [-0.5,\,0.5] \times [-0.3,\,0.3] \times [-0.5,\,0.5]$, 
    where $v(t) = [v_x(t),\, v_y(t),\, \omega_z(t)]$ is the measured velocity, and the commands $v^{\mathrm{desired}}(t)$ are randomly sampled within the specified ranges.

    \item \textbf{Early Termination Rate:} the percentage of episodes that ended before the evaluation horizon $T_{\mathrm{hor}}$ due to failure conditions. A lower rate indicates more reliable climbing. Early termination was triggered when any of the following occurred: (i) the robot detached from the wall and fell to the ground, or (ii) all feet remained attached for more than 5.0\,s indicating that the robot did not move for at least half of the episode.
    
    \item \textbf{Average Walking Time:} the mean duration that episodes lasted (up to $T_{\mathrm{hor}}$). Higher values indicate the policy can sustain crawling for longer without failure.
    
    \item \textbf{Retention:} the fraction of stance phase time during which magnetic adhesion force is actively applied. Retention quantifies how reliably the robot maintains magnetic contact during stance. A higher value indicates that the feet remain attached whenever required or promptly re-attach after a slip, thereby minimizing periods without support.
    
    \item \textbf{Recovery Rate:} the percentage of adhesion failures (caused by $prob_{attach} < 1$) that were successfully recovered within $\Delta T$ without episode termination. A higher rate indicates that the policy can better withstand adhesion disturbances.
\end{itemize}


\paragraph{Ablation Analysis on Training Components}
Table~\ref{tab:ablation_summary} summarizes the ablation study results, highlighting the contribution of curriculum learning, probabilistic adhesion, and adhesion modeling. Without curriculum learning, the robot tends to remain attached to the wall without detaching its feet, resulting in little actual movement and consequently a relatively low velocity tracking error compared to the case without adhesion modeling. However, this behavior results in extremely high early termination rates, since the robot fails to move forward and effectively remains stationary. Consequently, retention is artificially high because the feet stay attached throughout the episodes. In contrast, removing probabilistic adhesion shows performance comparable to the full model when $p_{\mathrm{attach}}=1.0$, but under $p_{\mathrm{attach}}=0.85$ the retention and recovery rates significantly degrade. This demonstrates that introducing stochastic adhesion improves robustness against attachment failures. Finally, training without adhesion modeling yields the worst performance across metrics, as the mismatch with the real magnetic mechanism leads to frequent detachment from the wall, large velocity tracking errors, and unstable climbing behavior.
\begin{table*}[t]
\caption{Ablation of curriculum and probabilistic adhesion in simulation. 
Metrics are computed over $T_{\mathrm{hor}} = 10\,\text{s}$ and $N = 100$ episodes}
\label{tab:ablation_summary}
\centering
\scriptsize   

\begin{tabular}{lccccc}
\multicolumn{6}{c}{\textbf{Results with $p_{\mathrm{attach}} = 1.0$}} \\
\toprule
\textbf{Condition} & \textbf{Vel.\ RMSE (mean~$\pm$~std).} & \textbf{Early Term.} &
\textbf{Avg.\ Time} & \textbf{Retention} \\
 & (m/s) & (\%) & (s) & (\%) & \\
\midrule
Full (ours)        & 0.1021~$\pm$~0.0890  & 0.00  & 10.0000~$\pm$~0.0000 & 79.67~$\pm$~8.00 \\
w/o Curriculum     & 0.2055~$\pm$~0.1803  & 88.00  & 5.8724~$\pm$~1.7620 & 98.50~$\pm$~9.26 \\
w/o Probabilistic  & 0.0844~$\pm$~0.0728  & 3.00  & 9.7639~$\pm$~1.3505 & 76.33~$\pm$~14.70 \\
w/o Modeling       & 5.1106~$\pm$~26.2067 & 48.00  & 6.6430~$\pm$~3.7370 & 44.36~$\pm$~35.53\\
\bottomrule
\end{tabular}

\vspace{0.75em}

\begin{tabular}{lccccc|ccc}
\multicolumn{9}{c}{\textbf{Results with $p_{\mathrm{attach}} = 0.85$}} \\
\toprule
\textbf{Condition} & \textbf{Vel.\ RMSE} & \textbf{Early Term.} &
\textbf{Avg.\ Time} & \textbf{Retention} & & 
\multicolumn{3}{c}{\textbf{Recovery Rate (\%)}} \\
 & (m/s) & (\%) & (s) & (\%) & & 
$\Delta T$=1.2s & $\Delta T$=2.4s & $\Delta T$=3.6s \\
\midrule
Full (ours)        & 0.5567~$\pm$~7.3437 & 4.00 & 9.7300~$\pm$~1.3699 & 72.93~$\pm$~18.46 & & 100.00~$\pm$~0.00 &   98.79~$\pm$~10.08 & 97.87~$\pm$~13.75 \\
w/o Probabilistic  & 6.4677~$\pm$~29.1904 & 66.00 & 6.5077~$\pm$~3.2063 & 42.51~$\pm$~32.31 & & 96.91~$\pm$~16.99 &   58.74~$\pm$~47.79 & 47.93~$\pm$~48.36 \\
\bottomrule
\end{tabular}
\end{table*}

\subsection{Hardware Validation}
\begin{figure}[t]
    \centering
    \includegraphics[width=\columnwidth]{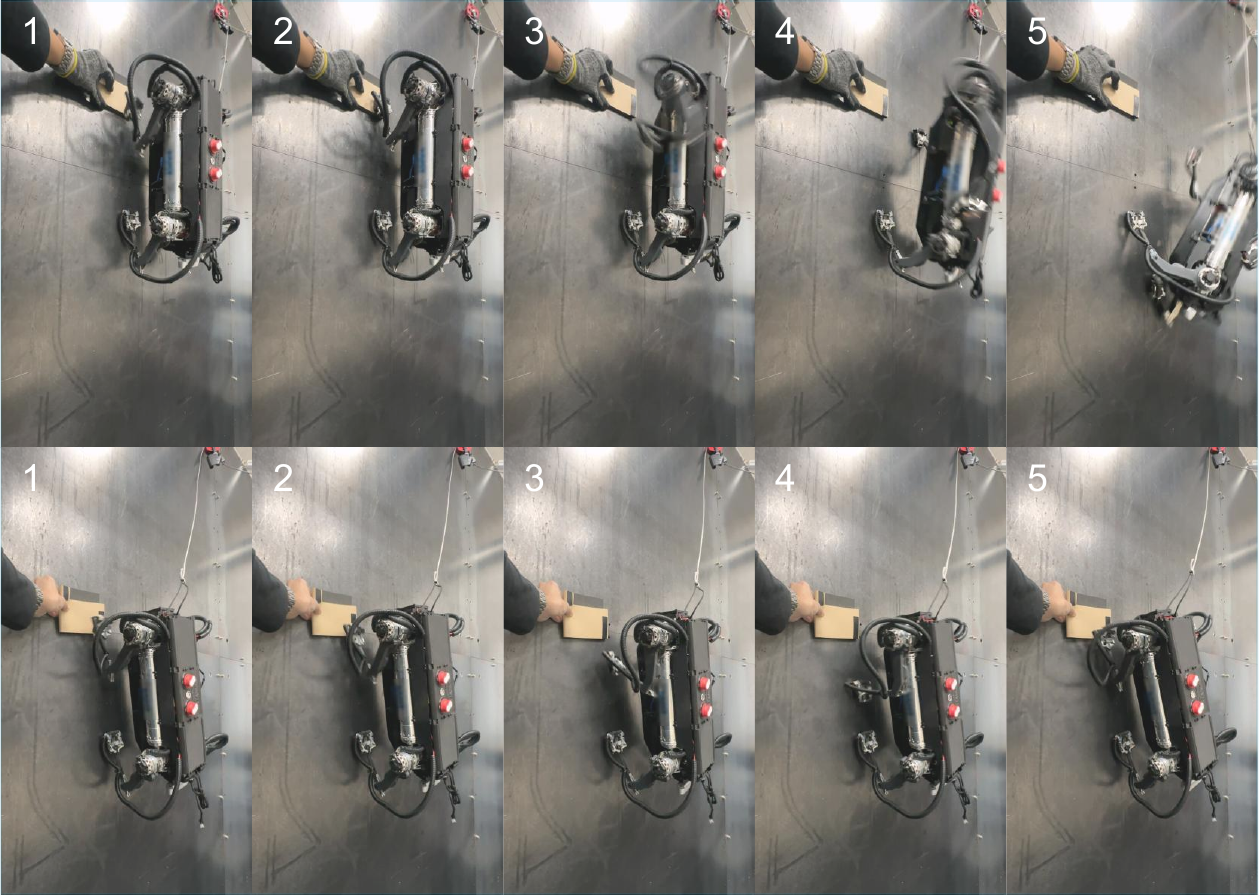}
    \caption{Snapshots from a hardware trial. (Top) MPC fails once adhesion is lost. (Bottom) Our RL policy recovers from a failed adhesion and resumes stable crawling.}
    \label{fig:hardware_seq}
\end{figure}

We further validate the proposed framework on the untethered quadrupedal magnetic climbing robot. 
In vertical wall experiments, the RL policy successfully traverses over non-ferromagnetic 
surfaces while recovering from induced adhesion failures. 
As a baseline, we compare against the Model Predictive Control (MPC) controller 
proposed in prior work~\cite{hong2022agile}, which assumes perfect adhesion. 
Snapshots in Fig.~\ref{fig:hardware_seq} show a representative hardware trial. 
When a front foot fails to attach, the RL policy shifts weight, reattempts adhesion, and resumes crawling. 
In contrast, the MPC baseline is unable to recover once adhesion fails, leading to immediate instability. 

\begin{figure}[t]
    \centering
    \includegraphics[width=\columnwidth]{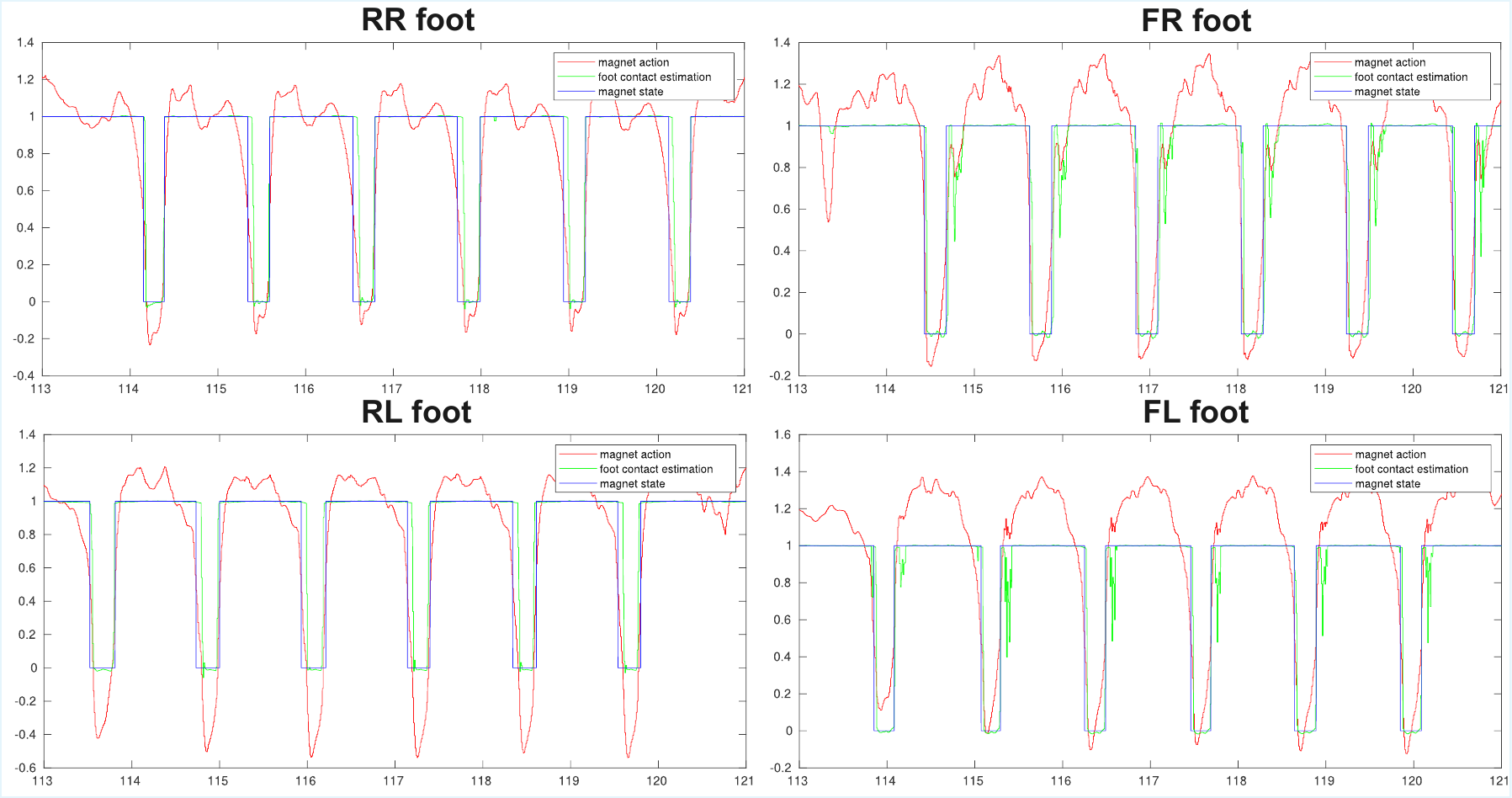}
    \caption{
    Logged signals from a hardware climbing trial. 
    For each leg (RR, FR, RL, FL), the plots show magnet activation command, foot contact estimation, and the measured EPM state, illustrating their synchronization during climbing.
    }
    \label{fig:hardware_log}
\end{figure}

In addition to visual demonstrations, we also log internal signals during hardware trials. Fig.~\ref{fig:hardware_log} illustrates the time traces of (i) the policy output for magnet activation, (ii) the estimated foot contact confidence, and (iii) the measured on/off state of the EPMs. While contact estimation occasionally fluctuated around the threshold due to mismatches with simulation, we adopted a simple rule in hardware: the magnet remained ON unless the contact probability stayed below 0.5 for more than 0.02 s. These traces confirm that the policy correctly synchronizes magnet actions with actual contact events, and that the EPMs are reliably switched in accordance with the commanded activation.

To further ensure reproducibility, multiple walking tests were conducted to verify that the learned controller can consistently achieve stable climbing locomotion, and the results are provided in the supplementary video.

\section{CONCLUSION AND FUTURE WORK}

We proposed a reinforcement learning framework for robust wall-climbing locomotion with the quadrupedal magnetic robot. The framework combines (i) a three-phase curriculum that transitions from ground crawling to vertical climbing, (ii) a realistic adhesion model of electropermanent-magnetic (EPM) feet that captures partial and failed contacts, and (iii) stochastic adhesion failures during training to promote recovery strategies. Through simulation and hardware experiments, the learned controller demonstrated stable vertical crawling, high adhesion retention, and effective slip recovery, while a model predictive control (MPC) baseline failed immediately under adhesion loss. These results highlight the benefits of learning-based control trained with explicit adhesion uncertainty.

Looking ahead, we plan to extend this framework to more diverse ferromagnetic environments, including curved surfaces, gaps, and irregular steel structures. We also aim to realize richer climbing skills, such as transitions across orientations (floor–wall–ceiling), enabling the robot to traverse a broader range of industrial and exploratory settings.

\addtolength{\textheight}{-0.5cm}   









\bibliographystyle{IEEEtran}
\bibliography{root}

\end{document}